\documentclass{article}

\usepackage{iclr2027_conference,times}

\usepackage{amsmath}
\usepackage{amssymb}
\usepackage{amsfonts}
\usepackage{amsthm}
\usepackage{mathtools}

\usepackage{graphicx}
\usepackage{booktabs}
\usepackage{xcolor}
\usepackage{microtype}
\usepackage{url}
\usepackage{hyperref}

\theoremstyle{plain}
\newtheorem{theorem}{Theorem}[section]
\newtheorem{proposition}[theorem]{Proposition}
\newtheorem{corollary}[theorem]{Corollary}

\theoremstyle{definition}

\theoremstyle{remark}
\newtheorem{remark}[theorem]{Remark}

\title{GIST: Gibbs-Targeted Hierarchical Data Selection for Offline RLVR}

\author{
Xuliang Huang\textsuperscript{1},
Tianxiang Zhao\textsuperscript{1},
Chengwei Qin\textsuperscript{1,2},
Yao Shu\textsuperscript{1,2}\thanks{Corresponding author.}\\
\textsuperscript{1}The Hong Kong University of Science and Technology (Guangzhou)\\
\textsuperscript{2}The Hong Kong University of Science and Technology
}

\iclrfinalcopy

\begin{document}

\maketitle

\pagestyle{plain}
\thispagestyle{plain}

\begin{abstract}

Offline reinforcement learning with verifiable rewards (RLVR) trains on a fixed pool of sampled responses and typically supervises every token of every response, although responses and tokens are not equally informative.
We study which ones to supervise through the Gibbs-optimal policy of KL-regularized RLVR, which offline training targets, and measure the mismatch between a policy and this target with a bidirectional KL divergence.
At the reference policy, this mismatch factorizes exactly and separates the two outcomes: correct responses contribute mainly through the forward term, concentrated on their low-probability tokens, and incorrect responses through the reverse term, concentrated on the responses the policy generates often and on their high-probability tokens.
Based on this analysis, we propose \textbf{GIST (Gibbs-Targeted Hierarchical Data Selection)}, a two-level selection method that approximates these principles with signals available before training.
GIST uses the sampling temperature to select correct responses on-policy and incorrect responses near the policy's mode, and it uses the initial policy's token probabilities to supervise only the tokens where the mismatch lies.
Applied to two offline training objectives, GIST raises the average accuracy over GSM8K, MATH-500, and MBPP for all three models under both objectives.
In a controlled comparison on Qwen3-0.6B/GSM8K, it outperforms seven selection baselines that choose from the same candidates while supervising 66.6\% fewer tokens than training on all candidates of the same prompts.
\end{abstract}


\begin{figure}[!h]
\centering
\begin{minipage}[c]{0.5\linewidth}
\centering
\includegraphics[width=\linewidth]{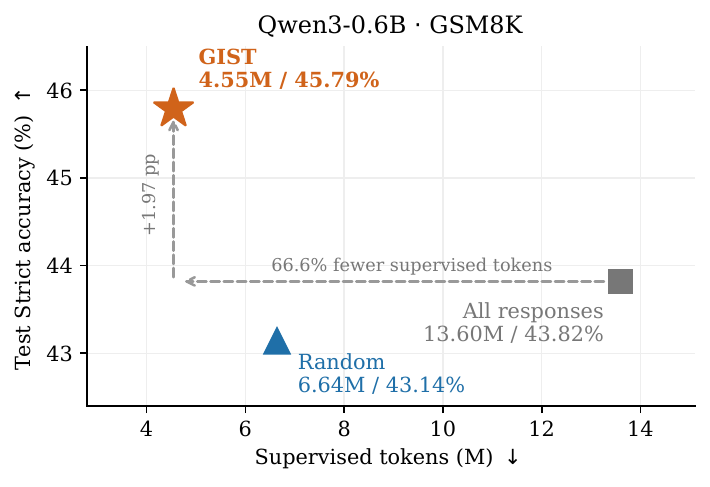}
\end{minipage}\hfill
\begin{minipage}[c]{0.43\linewidth}
\caption{Test Strict accuracy versus supervised tokens on Qwen3-0.6B/GSM8K. All methods use BOLT on the same candidate pool. GIST reaches the highest accuracy while supervising 66.6\% fewer tokens than training on all candidates and 31.5\% fewer than random selection.}
\label{fig:teaser}
\end{minipage}
\vspace{-0.5em}
\end{figure}

\section{Introduction}
\label{sec:intro}
Offline reinforcement learning with verifiable rewards (RLVR) trains a language model on a fixed pool of responses whose outcomes are checked automatically~\citep{lambert2024tulu3,shao2024deepseekmath,guo2025deepseekr1,wang2024oreo}.
Common approaches supervise every token of the responses they retain, although responses and tokens are not equally informative.
Deciding which responses to keep and which of their tokens to supervise is therefore a central design choice in offline RLVR.

Existing methods address parts of this choice: rejection sampling keeps verified correct responses~\citep{zelikman2022star,yuan2023rft}, reward-weighted objectives such as BOLT weight responses by their reward~\citep{peng2019awr,shu2026bolt}, and other methods subsample rollouts~\citep{li2026hes,xu2025pods} or select tokens by entropy or probability~\citep{wang2025highentropy,liu2026profit}.
They leave open how to combine response selection with outcome-dependent token supervision, although whether a token deserves supervision depends on both the outcome and how the response was sampled.
A low-probability token in a high-temperature sample may be noise, but in a verified correct response it is more likely a valid step that the policy under-weights, while failures sampled at low temperature reveal errors near the policy's mode.

We study selection through the Gibbs-optimal policy of KL-regularized RLVR, the target that offline training fits, and measure the mismatch between the policy and this target with a bidirectional KL divergence and a KL-curvature score that bounds it from above and below.
The forward term governs correct responses, whose informative instances are under-represented and whose informative tokens have low probability.
The reverse term governs incorrect responses, whose informative instances are over-represented and whose informative tokens have high probability.
At the reference policy, where offline selection takes place, the mismatch factorizes exactly.

We then propose \textbf{GIST} (Gibbs-Targeted Hierarchical Data Selection), a two-level selection method that approximates these principles with three quantities available at data-construction time: the sampling temperature, the verifier outcome, and the initial policy's token probabilities.
At the response level, GIST takes correct responses from on-policy samples and incorrect responses from near-mode samples, and it keeps the prompts that supply both.
At the token level, it supervises the low-probability tokens of correct responses and the high-probability tokens of incorrect ones.

Paired with two offline objectives, BOLT and SFT, GIST improves accuracy on GSM8K, MATH-500, and MBPP in nearly all settings (29 of 30) across three models from two families, and it raises the average accuracy in every case.
On Qwen3-0.6B/GSM8K, where all methods choose from the same candidates, GIST outperforms seven selection baselines while supervising 66.6\% fewer tokens and taking 49.9\% fewer optimizer steps than training on all candidates of the same prompts (Figure~\ref{fig:teaser}).

Our contributions are as follows.
(i)~We derive selection principles for offline RLVR from the bidirectional mismatch to the Gibbs-optimal target, separating the roles of correct and incorrect responses at the response and token levels (Section~\ref{sec:theory}).
(ii)~We propose GIST, a two-level selection method that approximates these principles from signals observed during data construction (Section~\ref{sec:methodology}).
(iii)~Across three models and two offline objectives, GIST improves nearly all benchmark results, and in a controlled comparison it outperforms seven selectors while supervising 66.6\% fewer tokens (Section~\ref{sec:experiments}).

\section{Preliminaries}
\label{sec:preliminaries}

\paragraph{KL-regularized RLVR.}
For a prompt $x$ and response $y$, a verifier assigns a binary reward $r(x,y)\in\{0,1\}$, with $r=1$ for correct and $r=0$ for incorrect responses.
Throughout, the reference policy $\pi_{\mathrm{ref}}$ is the initial policy $\pi_0$, and KL-regularized RLVR~\citep{ouyang2022instructgpt,lambert2024tulu3} maximizes
\begin{equation}
J_x(\pi)=\mathbb{E}_{y\sim\pi(\cdot\mid x)}\bigl[r(x,y)\bigr]-\beta\,\mathrm{KL}\bigl(\pi(\cdot\mid x)\,\|\,\pi_{\mathrm{ref}}(\cdot\mid x)\bigr),
\label{eq:rlvr-objective}
\end{equation}
where $\beta>0$ controls the strength of regularization.

\begin{proposition}[Gibbs target]
\label{prop:gibbs}
For a fixed prompt $x$, the unique maximizer of Eq.~\eqref{eq:rlvr-objective} is
\begin{equation}
\pi^\star(y\mid x)=\frac{\pi_{\mathrm{ref}}(y\mid x)\,e^{r(x,y)/\beta}}{Z(x)},
\qquad
Z(x)=\sum_{y'}\pi_{\mathrm{ref}}(y'\mid x)\,e^{r(x,y')/\beta},
\label{eq:gibbs-target}
\end{equation}
and for any policy $\pi$ with finite KL divergence to $\pi_{\mathrm{ref}}$,
\begin{equation}
J_x(\pi^\star)-J_x(\pi)=\beta\,\mathrm{KL}\bigl(\pi(\cdot\mid x)\,\|\,\pi^\star(\cdot\mid x)\bigr).
\label{eq:gap}
\end{equation}
\end{proposition}

The target reweights the reference policy by the exponentiated reward~\citep{peters2010reps,korbak2022rlkl,rafailov2023dpo}, and Eq.~\eqref{eq:gap} identifies the objective gap with the reverse KL to $\pi^\star$.

\paragraph{Offline fitting of the Gibbs target.}
Offline training fits $\pi^\star$ using a fixed pool of responses drawn from a rollout distribution $\mu$, with $\mu(y\mid x)>0$ wherever $\pi^\star(y\mid x)>0$.
Rewriting the forward KL as an expectation under $\mu$ gives
\begin{equation}
\mathbb{E}_x\bigl[\mathrm{KL}(\pi^\star\|\pi_\theta)\bigr]
=\mathbb{E}_x\,\mathbb{E}_{y\sim\mu}\bigl[-\rho(x,y)\log\pi_\theta(y\mid x)\bigr]+C,
\qquad
\rho(x,y)=\frac{\pi_{\mathrm{ref}}(y\mid x)}{\mu(y\mid x)}\,\frac{e^{r(x,y)/\beta}}{Z(x)},
\label{eq:bolt-weight}
\end{equation}
where $C$ does not depend on $\theta$.
When $\mu=\pi_{\mathrm{ref}}$, the weight reduces to $e^{r(x,y)/\beta}/Z(x)$, and Eq.~\eqref{eq:bolt-weight} becomes the reference-sampled weighted-likelihood objective of BOLT~\citep{shu2026bolt}.
This objective fits the target in the forward direction, whereas the objective gap in Eq.~\eqref{eq:gap} is a reverse KL.
Its weights also separate the two outcome classes: an incorrect response receives $e^{-1/\beta}$ times the weight of a correct response to the same prompt.
Section~\ref{sec:theory} uses both directions to locate the mismatch, and Section~\ref{sec:method-train} describes how the weights are computed in training.

\paragraph{Token-level view.}
Let $s_t=(x,y_{<t})$ denote the prefix at step $t$, and let $Z(s)$ be the expectation of $e^{r/\beta}$ over reference continuations of $s$, so that $Z(x)$ is the normalizer in Eq.~\eqref{eq:gibbs-target} and $Z(s)=e^{V(s)/\beta}$ for the soft value $V$~\citep{ziebart2010modeling,haarnoja2017soft,rafailov2024rq}.
The Gibbs target then factorizes over tokens as
\begin{equation}
\pi^\star(a\mid s)=\pi_{\mathrm{ref}}(a\mid s)\,\frac{Z(sa)}{Z(s)},
\label{eq:gibbs-token-conditional}
\end{equation}
so a token is more likely under the target than under the reference policy exactly when it raises the soft value.
For two distributions $p$ and $q$ over responses with conditionals $p_t=p(\cdot\mid s_t)$ and $q_t=q(\cdot\mid s_t)$, the chain rule for KL divergence~\citep{cover2006elements} gives
\begin{equation}
\mathrm{KL}(p\|q)=\sum_t\mathbb{E}_{s_t\sim p}\bigl[\mathrm{KL}(p_t\|q_t)\bigr].
\label{eq:token-kl-chain}
\end{equation}
Throughout, $\varphi(u)=u\log u-u+1\ge0$, so that $\mathrm{KL}(p\|q)=\sum_y q_y\,\varphi(p_y/q_y)$.

\section{A KL-Curvature View of Data Selection}
\label{sec:theory}

To decide which responses and tokens to supervise, we ask where the policy differs most from the Gibbs target of Section~\ref{sec:preliminaries}, first for whole responses and then for individual tokens (proofs in Appendix~\ref{app:theory}).

\subsection{Bidirectional Mismatch and Curvature Score}
\label{sec:theory-utility}

Fix a prompt $x$, and write $p_y=\pi^\star(y\mid x)$ and $q_y=\pi_\theta(y\mid x)$.
For $\lambda\ge0$, define
\begin{equation}
D_\lambda(p,q)=\mathrm{KL}(p\|q)+\lambda\,\mathrm{KL}(q\|p),
\qquad
U_y^\lambda=(p_y-q_y)^2\left(\frac{1}{q_y}+\frac{\lambda}{p_y}\right).
\label{eq:bidirectional-kl}
\end{equation}
The forward term is the target-fitting loss of Section~\ref{sec:preliminaries}; it measures how much the policy under-represents target behavior~\citep{minka2005divergence}.
The reverse term, scaled by $\beta$, is the RLVR objective gap (Proposition~\ref{prop:gibbs}); it measures how much policy mass lies on responses that the target down-weights.
The score $U_y^\lambda$ weights the squared probability gap by the local curvature of both KL terms~\citep{csiszar1967information}, so a gap of a given size counts more when it occurs at small probabilities.

\begin{theorem}[KL-curvature bounds]
\label{thm:kl-sandwich}
Let $p,q$ be positive on a common support with $\gamma\le p_y/q_y\le\Gamma$, where $0<\gamma\le1\le\Gamma$.
Then $c_-U^\lambda\le D_\lambda\le c_+U^\lambda$, where $U^\lambda=\sum_yU^\lambda_y$, $c_-=\tfrac12\min\{\Gamma^{-1},\gamma\}$, and $c_+=\tfrac12\max\{\gamma^{-1},\Gamma\}$.
As the density ratios converge uniformly to one, $D_\lambda=\tfrac12U^\lambda+o(U^\lambda)$.
\end{theorem}

Up to constants, the score thus bounds the mismatch from above and below, and it is exact to second order near the target.
The curvature factor also explains why a verifier label alone does not determine a response's value: the same probability gap counts more where the policy or the target places little mass.
The coefficient $\lambda$ weights the two directions and is used only in the analysis.

\subsection{Budgeted Response Selection}
\label{sec:theory-budget}

To select under a budget, we attribute the mismatch to individual responses through the contribution $K_y^\lambda=q_y\varphi(p_y/q_y)+\lambda p_y\varphi(q_y/p_y)$, which is nonnegative and satisfies $D_\lambda=\sum_yK^\lambda_y$ (Section~\ref{sec:preliminaries}).

\begin{corollary}[Budgeted selection guarantee]
\label{prop:coreset}
Let $S_U$ contain the $B$ candidates with the largest $U^\lambda_y$, and let $S^\star$ maximize $\sum_{y\in S}K^\lambda_y$ over subsets of size $B$.
Under the conditions of Theorem~\ref{thm:kl-sandwich}, $\sum_{y\in S_U}K_y^\lambda\ge(c_-/c_+)\sum_{y\in S^\star}K_y^\lambda$.
\end{corollary}

Selecting by the curvature score therefore captures a constant fraction of the largest attainable mismatch, similar to coreset guarantees~\citep{sener2018coreset,mirzasoleiman2020craig}, and this fraction approaches one as the policy approaches the target.
Note that the guarantee is about the mismatch captured at a fixed policy; it does not concern the outcome of training on the selected set.

With binary rewards, the target is $p_y=\pi_{\mathrm{ref}}(y\mid x)e^{r(x,y)/\beta}/Z(x)$, and the score behaves differently on the two classes.

\begin{corollary}[Informative correct and incorrect responses]
\label{cor:two-classes}
For fixed $\lambda>0$:
(i)~if $r=1$ and $q_y\ll p_y$, then $U^\lambda_y\sim p_y^2/q_y$ and the forward term dominates;
(ii)~if $r=0$ and $q_y\gg p_y$, then $U^\lambda_y\sim\lambda q_y^2/p_y$ and the reverse term dominates.
At comparable reference probabilities, (i) favors under-represented correct responses and (ii) favors over-represented incorrect responses.
\end{corollary}

Under-represented correct responses drive the forward (coverage) error, and over-represented incorrect ones drive the reverse error, which Proposition~\ref{prop:gibbs} identifies with the RLVR objective gap.
The two directions describe the same mismatch: mass moved onto under-represented correct responses must come from other responses, and the reverse term measures how much of it sits on over-represented errors.
Offline selection evaluates the mismatch at $q=\pi_0=\pi_{\mathrm{ref}}$, where both classes have an exact form.

\begin{proposition}[Exact mismatch at the reference policy]
\label{prop:ref-exact}
At $q=\pi_{\mathrm{ref}}$, $p_y/q_y=\kappa_r(x):=e^{r/\beta}/Z(x)$ and $K^\lambda_y=q_y\,h_\lambda(\kappa_r(x))$, where $h_\lambda(\kappa)=\kappa\log\kappa-\kappa+1+\lambda(\kappa-1-\log\kappa)$.
With $P(x)$ the reference success probability, $\kappa_1\approx1/P(x)>1$ and $\kappa_0\approx e^{-1/\beta}/P(x)<1$.
\end{proposition}

Hence every correct response is under-represented and every incorrect response is over-represented, as in Corollary~\ref{cor:two-classes}.
Within a class, $K^\lambda_y\propto q_y$, so frequently generated incorrect responses contribute most.
For correct responses, the under-representation is uniform at the response level and localized at the token level (Section~\ref{subsec:token-level-theory}).

\begin{remark}[The curvature score at the reference policy]
\label{rem:local}
At $q=\pi_{\mathrm{ref}}$, Eq.~\eqref{eq:bidirectional-kl} gives
\begin{equation}
U^\lambda_y=q_y\,(\kappa_r-1)^2\Bigl(1+\frac{\lambda}{\kappa_r}\Bigr),
\label{eq:u-ref}
\end{equation}
so $K^\lambda_y/U^\lambda_y$ is constant within a class and Corollary~\ref{prop:coreset} is exact there; across classes, Proposition~\ref{prop:ref-exact} gives the exact weights (Appendix~\ref{app:ref-policy}).
\end{remark}

\subsection{Token-Level Mismatch}
\label{subsec:token-level-theory}

Applying the chain rule of Eq.~\eqref{eq:token-kl-chain} to both terms of $D_\lambda$ decomposes the mismatch over prefixes, with the forward part weighted by $p(s_t)$ and the reverse part by $q(s_t)$.
For an observed token, we define its conditional score $U^\lambda_t$ and contribution $K^\lambda_t$ as in Eq.~\eqref{eq:bidirectional-kl} and Section~\ref{sec:theory-budget}, with $p_t,q_t$ evaluated at that token.
Theorem~\ref{thm:kl-sandwich} and Corollary~\ref{prop:coreset} also hold for these conditional quantities at each prefix, and the two regimes again separate.

\begin{corollary}[Informative tokens]
\label{cor:tokens}
For fixed $\lambda>0$ and comparable target conditionals across positions:
(i)~in a correct response, low-probability tokens ($q_t\ll p_t$) carry the forward mismatch, with $U^\lambda_t\sim p_t^2/q_t$;
(ii)~in an incorrect response, high-probability tokens ($q_t\gg p_t$) carry the reverse mismatch, with $U^\lambda_t\sim\lambda q_t^2/p_t$.
\end{corollary}

\begin{proposition}[Conditional token-level factorization]
\label{prop:token-exact}
At $\pi_\theta=\pi_{\mathrm{ref}}$, with $q_t=\pi_{\mathrm{ref}}(y_t\mid s_t)$ and $\rho_t=Z(s_ty_t)/Z(s_t)$, we have $K^\lambda_t=q_t\,h_\lambda(\rho_t)$, $\prod_t\rho_t=\kappa_{r(x,y)}(x)$, and $\rho_t\le1/q_t$.
\end{proposition}

By Eq.~\eqref{eq:gibbs-token-conditional}, $\rho_t$ is the token's soft-value ratio; GIST ranks positions by these conditional quantities and does not estimate the prefix weights (Appendix~\ref{app:theory}).
The bound $\rho_t\le1/q_t$ supports Corollary~\ref{cor:tokens}(i) at the reference policy even without the comparability condition: a large soft-value gain requires a low $q_t$, although a low probability alone does not imply a large gain.
For an incorrect response, $\log Z$ decreases by $\log(1/\kappa_0)\approx1/\beta+\log P(x)$ in total, and for a given decrease, $K^\lambda_t$ grows with $q_t$, as in (ii).
Because $h_\lambda$ is increasing on $[1,\infty)$, the largest conditional contribution attainable at a position with $\rho_t\ge1$ is
\begin{equation}
q_t\,h_\lambda(1/q_t)=\log\tfrac{1}{q_t}-1+q_t+\lambda\bigl(1-q_t-q_t\log\tfrac{1}{q_t}\bigr),
\label{eq:token-cap}
\end{equation}
which decreases in $q_t$.
The high-probability tokens of a correct response therefore account for little of its mismatch.

\begin{remark}[Where $\beta$ enters selection]
\label{rem:beta}
The within-class ranking by $q_y$, the bound $\rho_t\le1/q_t$, and the signs of the soft-value totals hold for every $\beta>0$, so the four principles in Table~\ref{tab:principles} do not depend on $\beta$.
Only the balance between the two classes depends on $\beta$ and $\lambda$, through the ratio of their total contributions,
\begin{equation}
R_\lambda(x)=\frac{\bigl(1-P(x)\bigr)\,h_\lambda(\kappa_0)}{P(x)\,h_\lambda(\kappa_1)}.
\label{eq:class-ratio}
\end{equation}
At $P(x)=0.5$ and $\beta=0.1$, $R_\lambda$ is 2.6 for $\lambda=0$ and 13.4 for $\lambda=1$; at $\beta=0.5$, it is 1.8 and 2.5.
Since the analysis does not fix the class balance, GIST sets it through the number of retained responses in each class.
\end{remark}

Table~\ref{tab:principles} summarizes the four principles, each of which follows from the curvature of the two KL terms.

\begin{table}[t]
\centering
\caption{Selection principles implied by the bidirectional mismatch.}
\label{tab:principles}
\small
\begin{tabular}{llll}
\toprule
Class & Dominant term & Response level & Token level \\
\midrule
Correct   & forward (coverage)        & under-represented & low-probability tokens \\
Incorrect & reverse (over-commitment) & over-represented  & high-probability tokens \\
\bottomrule
\end{tabular}
\end{table}

\section{GIST: Gibbs-Targeted Hierarchical Data Selection}
\label{sec:methodology}

\begin{figure}[t]
\centering
\includegraphics[width=\linewidth,trim=20 57 21 121,clip]{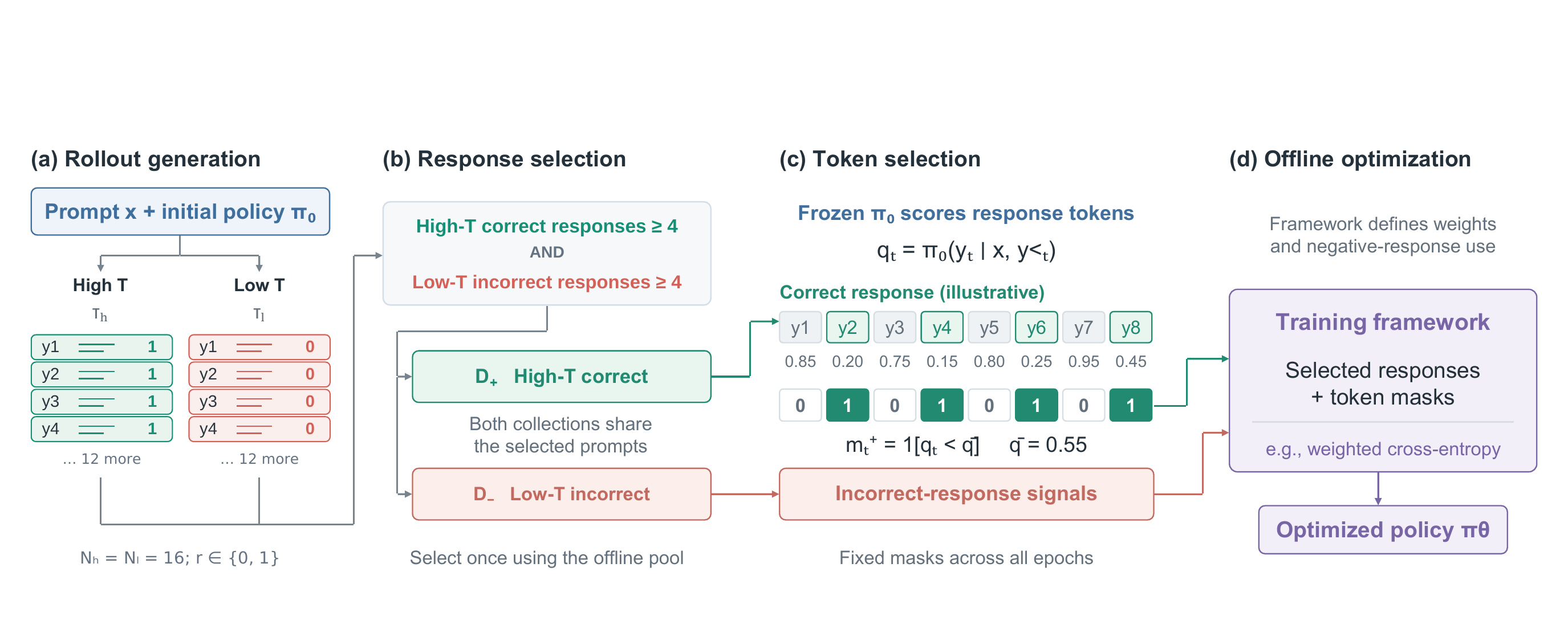}
\caption{Overview of GIST (schematic). (a)~Rollouts on-policy ($\tau_h=1$) and near the mode ($\tau_l$). (b)~Correct responses are taken from $\tau_h$ and incorrect ones from $\tau_l$. (c)~The frozen initial policy scores tokens to mask both classes. (d)~Selected responses and masks are passed to an offline training objective.}
\label{fig:method}
\end{figure}

The principles in Table~\ref{tab:principles} depend on target probabilities and soft values, which are not available during data construction.
Instead of estimating these quantities, GIST uses three observable signals: the sampling temperature, the verifier outcome, and the initial-policy token probability.
With these signals, it selects responses and then tokens within them, and it keeps the prompts that supply both outcomes (Figure~\ref{fig:method}).
All selections are made once, before training.

\subsection{Prompt and Response Selection}
\label{sec:method-response}

For each prompt $x$, the initial policy $\pi_0=\pi_{\mathrm{ref}}$ generates $N_h=16$ responses at temperature $\tau_h=1.0$ and $N_l=16$ at $\tau_l=0.3$, without top-$p$~\citep{holtzman2020curious} or top-$k$ truncation.
A verifier assigns $r(x,y)=1$ to correct and $r(x,y)=0$ to incorrect responses.
GIST takes correct responses from the on-policy pool and incorrect responses from the near-mode pool; we call this assignment \emph{temperature pairing}:
\begin{equation}
\mathcal{Y}_+(x)=\{y\in\mathcal{Y}_h(x): r(x,y)=1\},
\qquad
\mathcal{Y}_-(x)=\{y\in\mathcal{Y}_l(x): r(x,y)=0\}.
\label{eq:resp-sets}
\end{equation}
Within each class, the contribution is proportional to $q_y$ (Proposition~\ref{prop:ref-exact}), so sampling at $\tau_l$ concentrates on the incorrect responses that contribute most.
For correct responses, we choose $\tau_h=1$ with the token level in mind: on-policy responses keep the low-probability steps that the mask selects (Section~\ref{sec:exp-analysis}), and they need no temperature correction in Eq.~\eqref{eq:bolt-weight}.
The subsequent selection and masking, however, make the training loss an approximation to that objective.
A prompt is \emph{eligible} if $|\mathcal{Y}_+(x)|\ge B_+$ and $|\mathcal{Y}_-(x)|\ge B_-$.
The quotas are a practical choice and do not follow from the analysis.
They guarantee samples of both classes and bound the weights of correct responses; the cost is that the hardest prompts are excluded.
For each eligible prompt, GIST samples $B_+=4$ correct and $B_-=4$ incorrect responses uniformly without replacement, treating duplicates as distinct instances.
The analysis does not fix the balance between the classes (Remark~\ref{rem:beta}), and we use equal quotas.

\subsection{Token Selection}
\label{sec:method-token}

For a retained response $y$, let $q_t=\pi_0(y_t\mid x,y_{<t})$ and let $\bar q$ be its mean over the response positions, excluding EOS.
The masks are
\begin{equation}
m^+_t=\mathbf{1}\bigl[q_t<\bar q\bigr]\ \text{for correct responses},
\qquad
m^-_t=\mathbf{1}\bigl[q_t>\bar q\bigr]\ \text{for incorrect responses}.
\label{eq:masks}
\end{equation}
These masks implement the token-level principles, with the response mean as the threshold between low and high probability.
Correct responses keep their low-probability tokens, where the under-represented steps lie (Corollary~\ref{cor:tokens}(i), Proposition~\ref{prop:token-exact}), and incorrect responses keep their high-probability tokens, the confident errors of Corollary~\ref{cor:tokens}(ii).
Both masks are applied in every run that uses GIST's token selection.
They depend only on the verifier outcome and the initial policy, not on the training objective, and are computed once with the frozen initial policy and stay fixed during training.

\subsection{Training with the Selected Data}
\label{sec:method-train}
GIST outputs the selected responses, their rewards, and the token masks, so it can be combined with any offline objective that trains on weighted responses; we use two.
BOLT~\citep{shu2026bolt}, the weighted-likelihood objective of Section~\ref{sec:preliminaries}, weights each response by $e^{r/\beta}$, rescaled so that the weights of the prompt's 32 candidates average to one.
The weights are computed on the full candidate pool before selection and are not renormalized afterward.
A correct response therefore receives a weight of about $32/n_1$, where $n_1$ is the number of correct candidates, so prompts that the policy solves less often count more, and an incorrect response receives $e^{-1/\beta}$ times that weight.
SFT instead gives every retained response unit weight.
Both objectives average, over the minibatch, each response's weight times its mean negative log-likelihood on the masked positions; unmasked positions remain in the context but receive no loss.
\section{Experiments}
\label{sec:experiments}

\subsection{Main Results}
\label{sec:exp-main}

\paragraph{Setup.}
We train Qwen3-0.6B, Qwen3.5-2B, and Gemma-4-E2B-it~\citep{yang2025qwen3,qwen2026qwen35,gemma2026gemma4} and evaluate on GSM8K, MATH-500, and MBPP-full~\citep{cobbe2021gsm8k,hendrycks2021math,lightman2024verify,austin2021mbpp}.
For math, we report Strict accuracy, which requires the prescribed answer format, and Flexible accuracy (Flex.), which also credits correctly extracted answers; for code, we report pass@1.
We pair GIST with two objectives: BOLT ($\beta=0.1$) and SFT, which imitates every training response with unit weight.
All trained methods share 32 candidates per prompt (16 per temperature) and validation-based checkpoint selection; each objective is trained either on all candidates or on GIST's selected subset (+ GIST; details in Appendix~\ref{app:impl}).

\providecommand{\gain}[1]{\,{\scriptsize\textcolor[HTML]{B22222}{(#1)}}}
\newlength{\gistcol}
\newlength{\gistmeth}
\providecommand{\loss}[1]{\,{\scriptsize\textcolor[HTML]{7A7A7A}{(#1)}}}
\providecommand{\plusgist}{\hspace{0.6em}+ GIST}
\ifcsname rowcolor\endcsname
  \providecommand{\gistrow}{\rowcolor[HTML]{F2F2F2}\cellcolor{white}}
\else
  \providecommand{\gistrow}{}
\fi
\begin{table}[t]
\centering
\setlength{\abovecaptionskip}{4pt}
\setlength{\belowcaptionskip}{1.5pt}
\caption{Test results with BOLT and SFT (\%). Avg.: mean of GSM8K Flex., MATH-500 Flex., and MBPP-full pass@1. Bold marks the best score within each model and column. Parentheses: change vs.\ Initial.}
\label{tab:main}
\small
\renewcommand{\arraystretch}{0.92}
\setlength{\tabcolsep}{2.6pt}
\settowidth{\gistcol}{\textbf{42.20}\gain{+39.60}}
\settowidth{\gistmeth}{BOLT + GIST}
\begin{tabular}{llcccccc}
\toprule
 & & \multicolumn{2}{c}{GSM8K $\uparrow$} & \multicolumn{2}{c}{MATH-500 $\uparrow$} & MBPP-full $\uparrow$ & \\
\cmidrule(lr){3-4}\cmidrule(lr){5-6}\cmidrule(lr){7-7}
Model & Method & Strict & Flex. & Strict & Flex. & pass@1 & Avg. $\uparrow$ \\
\midrule
Qwen3-0.6B & Initial & 40.86 & 46.10 & 44.20 & 49.80 & 22.80 & 39.57 \\
 & SFT & 42.08\gain{+1.22} & 47.38\gain{+1.28} & 44.40\gain{+0.20} & 50.00\gain{+0.20} & 24.20\gain{+1.40} & 40.53\gain{+0.96} \\
\gistrow & \plusgist & 43.82\gain{+2.96} & 48.14\gain{+2.04} & 44.80\gain{+0.60} & 50.40\gain{+0.60} & 28.80\gain{+6.00} & 42.45\gain{+2.88} \\
 & BOLT & 43.29\gain{+2.43} & 47.46\gain{+1.36} & 44.60\gain{+0.40} & 50.40\gain{+0.60} & 26.60\gain{+3.80} & 41.49\gain{+1.92} \\
\gistrow & \plusgist & \textbf{45.79}\gain{+4.93} & \textbf{49.73}\gain{+3.63} & \textbf{45.40}\gain{+1.20} & \textbf{51.00}\gain{+1.20} & \textbf{31.40}\gain{+8.60} & \textbf{44.04}\gain{+4.47} \\
\midrule
Qwen3.5-2B & Initial & 54.51 & 55.19 & 46.00 & 55.00 & 37.40 & 49.20 \\
 & SFT & 55.04\gain{+0.53} & 55.80\gain{+0.61} & 46.20\gain{+0.20} & 55.00\gain{+0.00} & 39.60\gain{+2.20} & 50.13\gain{+0.93} \\
\gistrow & \plusgist & 56.18\gain{+1.67} & 56.71\gain{+1.52} & 46.80\gain{+0.80} & 55.80\gain{+0.80} & 41.20\gain{+3.80} & 51.24\gain{+2.04} \\
 & BOLT & 54.13\loss{$-$0.38} & 54.51\loss{$-$0.68} & 46.40\gain{+0.40} & 54.40\loss{$-$0.60} & \textbf{42.40}\gain{+5.00} & 50.44\gain{+1.24} \\
\gistrow & \plusgist & \textbf{56.63}\gain{+2.12} & \textbf{57.09}\gain{+1.90} & \textbf{47.40}\gain{+1.40} & \textbf{56.20}\gain{+1.20} & 41.80\gain{+4.40} & \textbf{51.70}\gain{+2.50} \\
\midrule
Gemma-4- & Initial & 12.28 & 26.99 & 2.60 & 38.60 & 55.00 & 40.20 \\
E2B-it & SFT & 12.81\gain{+0.53} & 27.82\gain{+0.83} & 12.40\gain{+9.80} & 42.00\gain{+3.40} & 55.80\gain{+0.80} & 41.87\gain{+1.67} \\
\gistrow & \plusgist & 13.87\gain{+1.59} & 29.04\gain{+2.05} & 41.80\gain{+39.20} & 53.00\gain{+14.40} & 56.80\gain{+1.80} & 46.28\gain{+6.08} \\
 & BOLT & 12.89\gain{+0.61} & 26.31\loss{$-$0.68} & 40.80\gain{+38.20} & 52.80\gain{+14.20} & 56.40\gain{+1.40} & 45.17\gain{+4.97} \\
\gistrow & \plusgist & \textbf{14.10}\gain{+1.82} & \textbf{29.57}\gain{+2.58} & \textbf{42.20}\gain{+39.60} & \textbf{53.20}\gain{+14.60} & \textbf{57.00}\gain{+2.00} & \textbf{46.59}\gain{+6.39} \\
\bottomrule
\end{tabular}
\vspace{-3pt}
\end{table}

\paragraph{Results.}
Averaged over the three benchmarks (Avg.\ in Table~\ref{tab:main}), GIST improves both objectives on every model, by 1.26--2.55 points over BOLT and 1.11--4.41 points over SFT, and BOLT + GIST has the highest average for all three models.
GIST improves over BOLT on both mathematical benchmarks for all three models under both metrics (Table~\ref{tab:main}).
On GSM8K, the gains reach 2.50 points on Strict and 3.26 on Flexible.
On Flexible, BOLT falls below the initial checkpoint for two models, whereas BOLT + GIST improves on it.
For Qwen3-0.6B, paired tests over the 1,319 GSM8K questions support both gains (Appendix~\ref{app:stats}).
BOLT also trains on prompts whose candidates are all incorrect (Appendix~\ref{app:impl}), so this compares the two pipelines end to end; Sections~\ref{sec:exp-selectors} and~\ref{sec:exp-ablation} isolate the selection components.
Gains on MATH-500 are smaller (0.40--1.80 points).
On MBPP-full, GIST gains 4.80 points on Qwen3-0.6B and 0.60 on Gemma-4-E2B-it, and trails BOLT by 0.60 on Qwen3.5-2B.

GIST also improves over SFT for every model, benchmark, and metric (Table~\ref{tab:main}).
The gains are 0.76--1.74 points on GSM8K, 0.40--0.80 on MATH-500 for the two Qwen models, and 1.00--4.60 on MBPP-full.
On Gemma-4-E2B-it, MATH-500 Strict rises from 12.40\% to 41.80\% while Flexible rises by 11.00 points, so most of this gain comes from following the prescribed answer format.
On GSM8K, the gains from GIST are similar under Flexible and Strict accuracy for both objectives, which suggests that they do not come only from answer formatting.

\subsection{Comparison with Selection Baselines}
\label{sec:exp-selectors}

All selectors share the candidate pool, the $1{,}593$ eligible prompts, and the BOLT objective (Table~\ref{tab:selectors}).
Random samples four correct and four incorrect responses uniformly and supervises all tokens, RFT-style keeps all correct responses, and the others adapt HES, PODS, ProFit, ETS, and RSI-S~\citep{li2026hes,xu2025pods,liu2026profit,wang2025highentropy,lv2026rsi} (Appendix~\ref{app:baselines}).
\begin{table}[t]
\centering
\setlength{\abovecaptionskip}{4pt}
\setlength{\belowcaptionskip}{1.5pt}
\caption{Selection methods with BOLT on a shared candidate pool (Qwen3-0.6B/GSM8K, \%).}
\label{tab:selectors}
\footnotesize
\renewcommand{\arraystretch}{0.92}
\setlength{\tabcolsep}{1.6pt}
\begin{tabular}{lcccccccc}
\toprule
 & Random & PODS & RFT-style & RSI-S & ETS & ProFit & HES & GIST \\
\midrule
Strict $\uparrow$ & 43.14 & 43.52 & 44.05 & 43.21 & 43.75 & 42.46 & 44.20 & \textbf{45.79} \\
Flex. $\uparrow$ & 48.60 & 48.22 & 48.14 & 47.84 & 47.54 & 46.47 & 47.31 & \textbf{49.73} \\
\bottomrule
\end{tabular}
\vspace{-3pt}
\end{table}

GIST achieves the highest accuracy on both metrics, 1.59--3.33 points above all seven baselines on Strict and 1.13--3.26 points on Flexible.
It exceeds Random by 2.65 points on Strict and 1.13 points on Flexible while supervising 31.5\% fewer tokens (Section~\ref{sec:exp-analysis}), and it exceeds training on all candidates by 1.97 and 3.03 points.
Appendix~\ref{app:stats} reports paired significance tests for these comparisons.

\subsection{Ablation Study}
\label{sec:exp-ablation}
\providecommand{\cmark}{\checkmark}
\providecommand{\xmark}{$\times$}

We ablate GIST on the shared pool, changing one design choice at a time (Tables~\ref{tab:components} and~\ref{tab:token-rule}).

\begin{table}[t]
\centering
\setlength{\abovecaptionskip}{4pt}
\setlength{\belowcaptionskip}{3pt}
\begin{minipage}[c]{0.5\linewidth}
\centering
\caption{Component ablation, Qwen3-0.6B on GSM8K (\%). \emph{Sub.}: 4+4 responses instead of all 32; \emph{Pair.}: correct from $\tau_h$, incorrect from $\tau_l$; \emph{Mask}: $m^+$ of Eq.~\eqref{eq:masks}; \emph{Wt.}: Boltzmann vs.\ unit. Parentheses: change vs.\ (f).}
\label{tab:components}
\small
\renewcommand{\arraystretch}{0.92}
\setlength{\tabcolsep}{2.4pt}
\begin{tabular}{ccccccc}
\toprule
 & Sub. & Pair. & Mask & Wt. & Strict $\uparrow$ & Flex. $\uparrow$ \\
\midrule
(a) & \xmark & \xmark & \xmark & \cmark & 43.82\loss{$-$1.97} & 46.70\loss{$-$3.03} \\
(b) & \cmark & \xmark & \xmark & \cmark & 43.14\loss{$-$2.65} & 48.60\loss{$-$1.13} \\
(c) & \cmark & \xmark & \cmark & \cmark & 43.59\loss{$-$2.20} & 48.75\loss{$-$0.98} \\
(d) & \cmark & \cmark & \xmark & \cmark & 43.75\loss{$-$2.04} & 47.61\loss{$-$2.12} \\
(e) & \cmark & \cmark & \cmark & \xmark & 43.82\loss{$-$1.97} & 48.14\loss{$-$1.59} \\
\midrule
(f) & \cmark & \cmark & \cmark & \cmark & \textbf{45.79} & \textbf{49.73} \\
\bottomrule
\end{tabular}
\end{minipage}\hfill
\begin{minipage}[c]{0.48\linewidth}
\centering
\caption{Token rule for correct responses, Qwen3-0.6B on GSM8K (\%). \emph{Kept}: approximate share of positions supervised per correct response (\%). $^\dagger$Same number of positions per response as GIST. Parentheses: change vs.\ GIST.}
\label{tab:token-rule}
\small
\renewcommand{\arraystretch}{0.92}
\setlength{\tabcolsep}{1.2pt}
\begin{tabular}{lccc}
\toprule
Positions & Kept & Strict $\uparrow$ & Flex. $\uparrow$ \\
\midrule
All tokens                  & 100 & 43.75\loss{$-$2.04} & 47.61\loss{$-$2.12} \\
Random$^\dagger$            & 31 & 44.81\loss{$-$0.98} & 45.64\loss{$-$4.09} \\
Highest-prob.$^\dagger$     & 31 & 44.05\loss{$-$1.74} & 47.23\loss{$-$2.50} \\
\midrule
Below-mean (GIST)           & 31 & \textbf{45.79} & \textbf{49.73} \\
\bottomrule
\end{tabular}
\end{minipage}

\vspace{8pt}
\begin{minipage}[c]{0.5\linewidth}
\centering
\makeatletter\def\@captype{figure}\makeatother
\includegraphics[width=\linewidth]{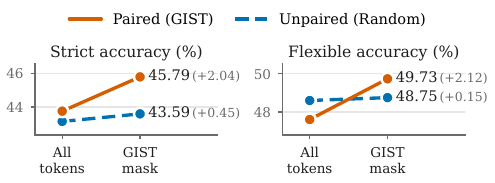}
\caption{Pairing and masking on Qwen3-0.6B/GSM8K. Unpaired responses are the Random selector; the mask helps mainly with paired responses.}
\label{fig:interaction}
\end{minipage}\hfill
\begin{minipage}[c]{0.48\linewidth}
\centering
\caption{Supervised tokens up to each method's selected checkpoint (Qwen3-0.6B/GSM8K). \emph{Reduction} is relative to all candidates.}
\label{tab:cost}
\small
\renewcommand{\arraystretch}{0.92}
\setlength{\tabcolsep}{1.5pt}
\begin{tabular}{lccc}
\toprule
 & All & Random & GIST \\
\midrule
Tokens            & 13.60M & 6.64M  & 4.55M \\
Reduction         & --     & 51.2\% & \textbf{66.6\%} \\
Strict $\uparrow$ & 43.82  & 43.14 & \textbf{45.79} \\
Flex. $\uparrow$  & 46.70  & 48.60 & \textbf{49.73} \\
\bottomrule
\end{tabular}
\end{minipage}
\vspace{-3pt}
\end{table}

Removing pairing (row~c), the mask (row~d), or the Boltzmann weights (row~e, the SFT + GIST entry of Table~\ref{tab:main}) from GIST lowers Strict/Flexible accuracy by 2.20/0.98, 2.04/2.12, and 1.97/1.59 points, respectively, and removing all selection (row~a) costs 1.97/3.03 (Table~\ref{tab:components}).
Replacing GIST's token rule also lowers both metrics (Table~\ref{tab:token-rule}): supervising all tokens costs 2.04 and 2.12 points, a random subset of the same size costs 0.98 and 4.09, and keeping the highest-probability tokens of correct responses costs 1.74 and 2.50.

Rows (b), (c), (d), and (f) of Table~\ref{tab:components} form a $2\times2$ design of pairing and masking (Figure~\ref{fig:interaction}).
The mask adds 2.04 and 2.12 points with pairing but only 0.45 and 0.15 without it, an interaction of 1.59 and 1.97 points.

\subsection{Discussion}
\label{sec:exp-analysis}

\paragraph{Token and response selection.}
The analysis in Section~\ref{sec:theory} holds at the policy that selects the data, so we compare its consequences with ablations that change one part of the data at a time.
The token-rule ablation keeps GIST's responses fixed and changes only which positions are supervised, which tests Corollary~\ref{cor:tokens}(i).
By Proposition~\ref{prop:token-exact}, a token can gain much soft value only if $q_t$ is low ($\rho_t\le1/q_t$), so the high-probability tokens of a correct response should carry little of its mismatch.
Keeping them instead of the low-probability ones loses most of the mask's benefit (Table~\ref{tab:token-rule}), and the ProFit adaptation, which also keeps high-probability tokens, is the weakest selector (Table~\ref{tab:selectors}).
Below-mean tokens make up about 31\% of correct responses at both temperatures, but their mean $q_t$ is 0.41 at $\tau_h$ and 0.58 at $\tau_l$ (Appendix~\ref{app:stats}); since the attainable contribution in Eq.~\eqref{eq:token-cap} decreases in $q_t$, the mask should find more informative tokens in on-policy responses.
Without the mask, a whole correct response contributes in proportion to $q_y$ (Proposition~\ref{prop:ref-exact}), which favors near-mode responses.
Together, these results predict that pairing and masking help mainly in combination, which matches the interaction in Section~\ref{sec:exp-ablation}.

\paragraph{Supervision and training cost.}
On the shared pool, GIST keeps 12,744 of the 50,976 candidates and supervises 66.6\% fewer tokens than training on all candidates and 31.5\% fewer than Random, while scoring higher than both (Table~\ref{tab:cost}, Figure~\ref{fig:teaser}, Appendix~\ref{app:cost}).
Masked positions still enter the forward and backward passes, so the compute saving over training on all candidates comes from keeping a quarter of the responses.
GIST reaches its selected checkpoint in 2,394 optimizer steps, about half of the 4,779 needed when training on all candidates, although its selected checkpoint comes after twice as many epochs.
Selection reuses BOLT's candidate pool and adds one forward pass of the frozen initial policy to compute $q_t$.

\paragraph{Generality.}
GIST selects data using only the sampling temperature, the verifier outcome, and the initial policy's token probabilities, with the same settings for every model and benchmark (Appendix~\ref{app:impl}).
Because the selection does not depend on the training objective, it can be paired with other offline methods.
SFT imitates every retained response with unit weight, including the incorrect ones, and SFT + GIST still matches or exceeds BOLT trained on all candidates in 14 of the 15 comparisons.
The benefit of the selection therefore does not come only from BOLT's reward weighting, although with GIST's data the weights add a further 0.20--2.60 points in every comparison.
\section{Related Work}
\label{sec:related_work}
 
\paragraph{Offline RLVR and policy optimization.}
Reinforcement learning with verifiable rewards (RLVR) improves reasoning using automatically checkable outcomes~\citep{lambert2024tulu3,shao2024deepseekmath,guo2025deepseekr1,yu2025dapo}.
Most RLVR methods optimize online with PPO-style updates~\citep{schulman2017ppo,liu2025drgrpo,yue2025does}, whereas offline approaches~\citep{levine2020offline} such as ILQL~\citep{snell2022ilql} and OREO~\citep{wang2024oreo} learn from previously collected rollouts.
BOLT~\citep{shu2026bolt} shows that Boltzmann-weighted likelihood, in the spirit of reward-weighted regression~\citep{peters2007rwr,peng2019awr}, on reference-policy samples fits the Gibbs-optimal target of KL-regularized RLVR.
These methods define the training objective; GIST instead decides which responses and tokens of a finite rollout pool to train on, and can be paired with BOLT or other offline objectives.
Work on alignment and distillation chooses which KL direction or $f$-divergence the model should minimize~\citep{go2023aligning,wang2024beyond,gu2024minillm}, whereas we use both directions to select the training data.
 
\paragraph{Data selection for reasoning post-training.}
Selection for post-training~\citep{zhou2023lima,xia2024less} has been studied at two granularities.
At the response level, rejection-sampling fine-tuning and self-training~\citep{zelikman2022star,yuan2023rft,dong2023raft,gulcehre2023rest,singh2024restem} keep verified correct trajectories, HES~\citep{li2026hes} ranks trajectories by the entropy of a few high-entropy tokens, PODS~\citep{xu2025pods} subsamples rollouts for policy updates, and dynamic sampling in DAPO~\citep{yu2025dapo} discards prompts whose rollouts are all correct or all incorrect; other methods filter RL prompts by difficulty or learning impact~\citep{bae2025online,li2025limr,wang2025oneshot}.
At the token level, Rho-1~\citep{lin2024rho1} and Token Cleaning~\citep{pang2025tokencleaning} select tokens by excess loss or update influence, and critical-token estimation~\citep{lin2024critical} contrasts models trained on correct and incorrect trajectories.
In RLVR, policy entropy governs exploration~\citep{cui2025entropy} and high-entropy forking tokens concentrate effective updates, which motivates entropy-based token selection (ETS)~\citep{wang2025highentropy}; RSI-S~\citep{lv2026rsi} and GMTS~\citep{lv2026gmts} use surprisal- and gradient-based token scores, and ProFit~\citep{liu2026profit} keeps high-probability tokens during SFT.
Each of these methods selects at a single level with its own criterion.
GIST derives both from the same bidirectional mismatch and applies them hierarchically, keeping, unlike ProFit, the low-probability tokens of correct responses.
\section{Conclusion}
\label{sec:conclusion}

We studied data selection for offline RLVR through the bidirectional mismatch between the policy and the Gibbs-optimal target.
In the regimes we identify, the forward term governs correct responses and the reverse term governs incorrect ones.
GIST turns this view into response- and token-level selection rules that use only the sampling temperature, the verifier outcome, and the initial policy's token probabilities.
Across three models and two offline objectives, it improves nearly all benchmark results, and it outperforms seven selectors on a shared candidate pool while supervising far fewer tokens than training on all candidates.
More broadly, the analysis suggests that which responses and which tokens to train on can be read from the policy itself, and we expect this view to apply to other offline objectives and to settings in which the data are re-selected as the policy changes.

 
\section*{Ethics Statement}
This work uses publicly available models and benchmarks and involves no human subjects or personal data.
We do not foresee ethical concerns beyond those of the underlying models.
 
\section*{Reproducibility Statement}
Proofs of all theoretical results are given in Appendix~\ref{app:theory}.
Appendix~\ref{app:impl} specifies the data construction and eligibility rule (Appendix~\ref{app:data}), the token masks (Appendix~\ref{app:masks}), the training objectives and hyperparameters (Appendix~\ref{app:objective}), the evaluation protocol (Appendix~\ref{app:eval}), and the paired tests (Appendix~\ref{app:stats}); Appendix~\ref{app:baselines} describes every selection baseline and ablation control.
All models and benchmarks are publicly available.
 
\section*{AI Use Statement}
In this work, we used generative AI tools to edit the paper for readability, to polish the formatting of tables and figures, and to check that the numbers reported in the text match the tables.
We have not used generative AI tools to develop the theoretical framework, formulate or prove the mathematical claims, design or implement the method and experiments, generate or process data, or interpret the results.
We have reviewed all AI-assisted edits, checked every reported number against our experimental logs, and take full responsibility for the content of the paper.

\bibliography{references}
\bibliographystyle{iclr2027_conference}

\clearpage
\appendix

\appendix

\section{Proofs}
\label{app:theory}

Throughout, $\varphi(u)=u\log u-u+1$, which satisfies $\varphi\ge0$, $\varphi(1)=\varphi'(1)=0$, and $\varphi''(u)=1/u$.

\subsection{Gibbs Target and Offline Fitting}

\begin{proof}[Proof of Proposition~\ref{prop:gibbs}]
Fix $x$ and write $p=\pi^\star(\cdot\mid x)$.
Since $r=\beta\log(p/\pi_{\mathrm{ref}})+\beta\log Z(x)$, for any $\pi$ with finite KL divergence to $\pi_{\mathrm{ref}}$,
\[
J_x(\pi)=\mathbb{E}_{\pi}\Bigl[\beta\log\frac{p}{\pi_{\mathrm{ref}}}\Bigr]+\beta\log Z(x)-\beta\,\mathbb{E}_{\pi}\Bigl[\log\frac{\pi}{\pi_{\mathrm{ref}}}\Bigr]
=\beta\log Z(x)-\beta\,\mathrm{KL}(\pi\|p).
\]
Nonnegativity of KL shows that $p$ is the unique maximizer with $J_x(p)=\beta\log Z(x)$, and subtracting gives Eq.~\eqref{eq:gap}.
\end{proof}

\paragraph{Derivation of Eq.~\eqref{eq:bolt-weight}.}
$\mathrm{KL}(p\|\pi_\theta)=\sum_y p_y\log p_y-\sum_y p_y\log\pi_\theta(y\mid x)$.
The first term does not depend on $\theta$.
If $\mu(y\mid x)>0$ wherever $p_y>0$, the second term equals $\mathbb{E}_{y\sim\mu}[-(p_y/\mu_y)\log\pi_\theta(y\mid x)]$, and $p_y/\mu_y=\rho(x,y)$ by Eq.~\eqref{eq:gibbs-target}.

\subsection{KL-Curvature Bounds and Budgeted Selection}

\begin{proof}[Proof of Theorem~\ref{thm:kl-sandwich}]
Because $\sum_y(p_y-q_y)=0$, we have $\mathrm{KL}(p\|q)=\sum_y q_y\varphi(p_y/q_y)$ and $\mathrm{KL}(q\|p)=\sum_y p_y\varphi(q_y/p_y)$.
For $u$ in an interval $[m,M]$ with $0<m\le1\le M$, Taylor's theorem around $u=1$ gives $(u-1)^2/(2M)\le\varphi(u)\le(u-1)^2/(2m)$.
With $u=p_y/q_y\in[\gamma,\Gamma]$,
\[
\frac{(p_y-q_y)^2}{2\Gamma q_y}\le q_y\varphi\Bigl(\frac{p_y}{q_y}\Bigr)\le\frac{(p_y-q_y)^2}{2\gamma q_y},
\]
and with $u=q_y/p_y\in[\Gamma^{-1},\gamma^{-1}]$,
\[
\frac{\gamma(p_y-q_y)^2}{2p_y}\le p_y\varphi\Bigl(\frac{q_y}{p_y}\Bigr)\le\frac{\Gamma(p_y-q_y)^2}{2p_y}.
\]
Multiplying the second pair by $\lambda$ and adding gives $c_-U^\lambda_y\le K^\lambda_y\le c_+U^\lambda_y$ for each $y$, and summing over $y$ gives the theorem.
For the local statement, $\varphi(1+h)=\tfrac12h^2+O(|h|^3)$ applied to both ratios gives $D_\lambda=\tfrac12U^\lambda+O(\delta\,U^\lambda)$ with $\delta=\sup_y|p_y/q_y-1|\to0$.
\end{proof}

\begin{proof}[Proof of Corollary~\ref{prop:coreset}]
Using the per-response bounds and the optimality of $S_U$ for $U^\lambda$,
$\sum_{S_U}K^\lambda_y\ge c_-\sum_{S_U}U^\lambda_y\ge c_-\sum_{S^\star}U^\lambda_y\ge(c_-/c_+)\sum_{S^\star}K^\lambda_y$.
\end{proof}

\begin{proof}[Proof of Corollary~\ref{cor:two-classes}]
Write $U^\lambda_y=(p_y-q_y)^2/q_y+\lambda(p_y-q_y)^2/p_y$.
(i)~With $a=q_y/p_y\to0$, the first term is $p_y^2(1-a)^2/q_y$ and the second is $\lambda p_y(1-a)^2$; their ratio is $1/(\lambda a)\to\infty$, so $U^\lambda_y\sim p_y^2/q_y$.
(ii)~With $b=p_y/q_y\to0$, the second term is $\lambda q_y^2(1-b)^2/p_y$ and the first is $q_y(1-b)^2$; their ratio is $\lambda/b\to\infty$, so $U^\lambda_y\sim\lambda q_y^2/p_y$.
At fixed target probability, the first expression decreases and the second increases in $q_y$.
\end{proof}

\subsection{Selection at the Reference Policy}
\label{app:ref-policy}

\begin{proof}[Proof of Proposition~\ref{prop:ref-exact}]
By Eq.~\eqref{eq:gibbs-target}, $p_y=q_y e^{r/\beta}/Z(x)=\kappa_r q_y$ when $q=\pi_{\mathrm{ref}}$.
Substituting into $K^\lambda_y=q_y\varphi(p_y/q_y)+\lambda p_y\varphi(q_y/p_y)$ gives $q_y\varphi(\kappa_r)+\lambda\kappa_rq_y\varphi(1/\kappa_r)$, and $\kappa\varphi(1/\kappa)=\kappa-1-\log\kappa$ yields $K^\lambda_y=q_y h_\lambda(\kappa_r)$.
With $r\in\{0,1\}$, $Z(x)=e^{1/\beta}P(x)+1-P(x)=e^{1/\beta}P(x)\bigl(1+O(e^{-1/\beta}/P(x))\bigr)$, so $\kappa_1=e^{1/\beta}/Z(x)\approx1/P(x)$ and $\kappa_0=1/Z(x)\approx e^{-1/\beta}/P(x)$.
\end{proof}

\paragraph{Between-class constants (Remark~\ref{rem:local}).}
At $q=\pi_{\mathrm{ref}}$, the density ratios range from $\gamma=\kappa_0$ to $\Gamma=\kappa_1$, so $c_-/c_+=\gamma^2\approx e^{-2/\beta}/P(x)^2$, below $10^{-7}$ at $\beta=0.1$ for $P(x)\ge0.25$.
For $\lambda=1$ and $P(x)\in[0.25,0.75]$, Eq.~\eqref{eq:u-ref} gives an incorrect-to-correct score ratio between about $5\times10^{2}$ and $8.5\times10^{4}$ for responses of equal probability, whereas the exact ratio $h_\lambda(\kappa_0)/h_\lambda(\kappa_1)$ is between 2 and 101.

\subsection{Token-Level Results}

\paragraph{Chain rule and target conditional.}
With a finite horizon and terminated sequences padded by an absorbing state, $\log(p(y\mid x)/q(y\mid x))=\sum_t\log(p_t(y_t\mid s_t)/q_t(y_t\mid s_t))$.
Taking expectations under $p$ and conditioning each term on its prefix gives the forward part of Eq.~\eqref{eq:token-kl-chain}; exchanging $p$ and $q$ gives the reverse part.
For the conditional, summing the Gibbs target over completions of a prefix gives $p(s)=\pi_{\mathrm{ref}}(s)Z(s)/Z(x)$, and the ratio of $p(sa)$ to $p(s)$ is Eq.~\eqref{eq:gibbs-token-conditional}.
Applying the proof of Theorem~\ref{thm:kl-sandwich} to $(p_t(y_t\mid s_t),q_t(y_t\mid s_t))$ gives $c_-U^\lambda_t\le K^\lambda_t\le c_+U^\lambda_t$, and the proofs of Corollaries~\ref{prop:coreset} and~\ref{cor:two-classes} apply verbatim to tokens, which proves Corollary~\ref{cor:tokens}.

\begin{proof}[Proof of Proposition~\ref{prop:token-exact}]
At $\pi_\theta=\pi_{\mathrm{ref}}$, Eq.~\eqref{eq:gibbs-token-conditional} gives $p_t/q_t=\rho_t$, and the computation in the proof of Proposition~\ref{prop:ref-exact} gives $K^\lambda_t=q_th_\lambda(\rho_t)$.
Since $Z(s_1)=Z(x)$ and $Z(s_{T+1})=e^{r(x,y)/\beta}$, the product $\prod_t\rho_t$ telescopes to $\kappa_{r(x,y)}(x)$.
Finally, $Z(s_t)=\sum_a\pi_{\mathrm{ref}}(a\mid s_t)Z(s_ta)\ge q_tZ(s_ty_t)$, so $\rho_t\le1/q_t$.
For the prefix-weighted share, the forward part $p(s_t)\,q_t\varphi(\rho_t)$ uses $p(s_t)=q(s_t)Z(s_t)/Z(x)$, and the reverse part $\lambda q(s_t)\,p_t\varphi(q_t/p_t)$ uses $\rho\,\varphi(1/\rho)=\rho-1-\log\rho$ with $p_t=\rho_tq_t$.
\end{proof}

\paragraph{Eq.~\eqref{eq:token-cap}.}
$h_\lambda'(\kappa)=\log\kappa+\lambda(1-1/\kappa)\ge0$ for $\kappa\ge1$, so $K^\lambda_t=q_th_\lambda(\rho_t)\le q_th_\lambda(1/q_t)$ whenever $1\le\rho_t\le1/q_t$.
Expanding gives the stated expression, whose derivative in $q_t$ is $(1-1/q_t)+\lambda\log q_t<0$ on $(0,1)$.
Along a correct response $\sum_t\log\rho_t=\log\kappa_1\approx\log(1/P(x))$, and along an incorrect response $\sum_t\log\rho_t=\log\kappa_0\approx-1/\beta-\log P(x)$.

\paragraph{Eq.~\eqref{eq:class-ratio}.}
By Proposition~\ref{prop:ref-exact}, the total contribution of class $r$ is $h_\lambda(\kappa_r)\sum_{y:r(x,y)=r}q_y$, which equals $P(x)h_\lambda(\kappa_1)$ for $r=1$ and $(1-P(x))h_\lambda(\kappa_0)$ for $r=0$.
The values quoted in Remark~\ref{rem:beta} follow by direct evaluation.

\section{Implementation Details}
\label{app:impl}

\subsection{Data Construction}
\label{app:data}

For each training prompt, the initial policy $\pi_0$ generates $N_l=16$ responses at $\tau_l=0.3$ and $N_h=16$ at $\tau_h=1.0$, without top-$p$ or top-$k$ truncation.
Rollouts use the chat template with thinking disabled, seed 42, and a stop at the closing \texttt{</answer>} tag, with at most 1,024 new tokens for GSM8K and MATH and 2,048 for MBPP.
Validation sets are 500 held-out training questions for GSM8K and for MATH, and MBPP task IDs 511--600.
The verifier assigns $r=1$ to correct and $r=0$ to incorrect responses: final-answer matching for GSM8K and MATH-500, and the unit tests for MBPP-full.
Malformed or truncated responses are kept and verified normally; a response from which no answer can be extracted or verified receives $r=0$.
Duplicate responses are kept as separate instances.
A prompt is eligible if it has at least $B_+=4$ correct $\tau_h$ responses and at least $B_-=4$ incorrect $\tau_l$ responses, and GIST samples four of each uniformly without replacement.
Table~\ref{tab:eligible} reports the eligible prompts; on Qwen3-0.6B/GSM8K, GIST retains 12,744 of the 50,976 candidates of the 1,593 eligible prompts.
In Table~\ref{tab:main}, BOLT and SFT train on all 32 candidates of every training prompt, and each objective with + GIST trains on GIST's selected subset.
The controlled comparisons of Sections~\ref{sec:exp-selectors}--\ref{sec:exp-analysis} use only the eligible prompts, and ``all candidates'' there refers to the 32 candidates of each eligible prompt.

\begin{table}[h]
\centering
\caption{Eligible prompts over training prompts.}
\label{tab:eligible}
\small
\begin{tabular}{lccc}
\toprule
Model & GSM8K & MATH & MBPP \\
\midrule
Qwen3-0.6B     & 1,593 / 6,973 & 1,619 / 7,000 & 66 / 374 \\
Qwen3.5-2B     & 2,574 / 6,973 & 1,100 / 7,000 & 65 / 374 \\
Gemma-4-E2B-it & 1,373 / 6,973 & 1,646 / 7,000 & 58 / 374 \\
\bottomrule
\end{tabular}
\end{table}

\subsection{Token Masks}
\label{app:masks}

Token probabilities $q_t=\pi_0(y_t\mid x,y_{<t})$ are computed once with the frozen initial policy and without temperature scaling, and the masks stay fixed during training.
The mean $\bar q$ is taken over the eligible response positions, which exclude prompt, padding, and EOS positions, so EOS never enters $\bar q$ or the mask.
Comparisons are strict, so positions equal to the mean are not selected.
Whenever GIST's token selection is applied, correct responses use $m^+_t=\mathbf{1}[q_t<\bar q]$ and incorrect responses use $m^-_t=\mathbf{1}[q_t>\bar q]$; the baselines keep their own token rules (Appendix~\ref{app:baselines}).
Unselected positions remain in the context but receive no loss.

\subsection{Training Objective}
\label{app:objective}

BOLT weights are computed within each prompt's full pool of 32 candidates before selection, as described in Section~\ref{sec:method-train}, with $r\in\{0,1\}$ and $\beta=0.1$, and are not renormalized after selection.
An incorrect response therefore has weight $e^{-10}\approx4.5\times10^{-5}$ times that of a correct response to the same prompt.
No off-policy correction is applied to responses from the $\tau_l$ pool, so $32/n_1$ estimates the inverse success rate of the mixed-temperature pool rather than $1/P(x)$; the BOLT baseline uses the same mixed pool.
For a prompt whose 32 candidates are all incorrect, every weight equals one, so BOLT imitates these responses; GIST excludes such prompts through eligibility.
The loss multiplies each response's mean masked token negative log-likelihood by its weight and averages over the minibatch.
SFT uses the same loss with $w=1$ for every response.
SFT + GIST, which is also row~(e) of Table~\ref{tab:components}, applies GIST's selection and masks, so the positions of incorrect responses that $m^-$ selects are imitated with the same weight as the positions of correct ones.
All other settings match BOLT.
We fine-tune LoRA adapters (rank 8, $\alpha=64$, dropout 0.05) with fused AdamW ($\beta_1=0.9$, $\beta_2=0.95$, weight decay 0.1), a constant schedule with 10\% warmup, BF16, and gradient checkpointing.
Learning rates are $5\times10^{-6}$ for GSM8K and $10^{-5}$ for MATH and MBPP; the global batch size is 32 for GSM8K and 128 for MATH and MBPP, and the maximum training length is 1,024 and 4,096 tokens, respectively.
Training runs for up to 12 epochs with early stopping (patience 2) after at least 5 epochs, and the checkpoint with the best validation Strict accuracy (pass@1 for MBPP) is evaluated on the test set.

\subsection{Evaluation}
\label{app:eval}

GSM8K is evaluated with lm-evaluation-harness~\citep{gao2024lmeval} using 8-shot Q:/A: prompts without the chat template and at most 256 new tokens.
GSM8K Strict requires ``The answer is \ldots'' followed by an exact match, and MATH Strict requires an exact match of the normalized last boxed answer; Flexible accuracy extracts the final answer from other formats.
MBPP-full reports pass@1, executing the benchmark's test cases in a bubblewrap sandbox with a 5-second timeout and CPU, memory, and file-size limits.
Each entry is a single training run.
The binomial standard error of a single test accuracy is about 1.4 points on GSM8K ($n=1{,}319$) and 2.2 points on MATH-500 and MBPP-full ($n=500$); Appendix~\ref{app:stats} reports paired tests for the main GSM8K comparisons.

\subsection{Paired Tests and Token Statistics}
\label{app:stats}

Table~\ref{tab:paired} reports paired tests on the 1,319 GSM8K test questions (Qwen3-0.6B), using per-question correctness of the two compared checkpoints: a paired bootstrap 95\% interval for the accuracy difference~\citep{efron1993bootstrap} and McNemar's test~\citep{mcnemar1947}, following \citet{dror2018hitchhiker}.

\begin{table}[h]
\centering
\caption{Paired tests on GSM8K (Qwen3-0.6B, 1,319 questions). Differences are computed from per-question counts and may differ by 0.01 from differences of the rounded accuracies in the main text.}
\label{tab:paired}
\small
\begin{tabular}{llccc}
\toprule
Comparison & Metric & Difference (pp) & Bootstrap 95\% CI & McNemar $p$ \\
\midrule
GIST $-$ BOLT          & Strict & $+2.50$ & $[+0.30,+4.62]$ & 0.027 \\
                       & Flex.  & $+2.27$ & $[+0.15,+4.40]$ & 0.044 \\
GIST $-$ Random         & Strict & $+2.65$ & $[+0.61,+4.70]$ & 0.014 \\
                        & Flex.  & $+1.14$ & $[-0.83,+3.11]$ & 0.293 \\
GIST $-$ All candidates & Strict & $+1.97$ & $[-0.08,+4.02]$ & 0.072 \\
                       & Flex.  & $+3.03$ & $[+0.99,+5.08]$ & 0.005 \\
\bottomrule
\end{tabular}
\end{table}

\paragraph{Below-mean tokens by sampling temperature.}
We measure the fraction of positions with $q_t<\bar q$ in correct responses, excluding EOS.
Pooled over tokens, it is 31.57\% at $\tau_l$ and 31.93\% at $\tau_h$ over all training prompts, and 31.38\% and 31.75\% over the 1,593 eligible prompts.
With each response weighted equally, it is 31.39\% and 30.69\% over all prompts, and 31.22\% and 30.56\% over eligible prompts.
Comparing correct responses within the same prompt, the high-temperature fraction is lower by 0.36 points over all prompts (95\% CI $[-0.50,-0.21]$) and by 0.57 points over eligible prompts ($[-0.74,-0.41]$).
Both temperatures thus contain a substantial fraction of below-mean tokens, and the sign of the small difference depends on aggregation, reflecting differences in response length.
The probability of the selected tokens differs more: on the eligible prompts, below-mean tokens of correct responses have mean (median) $q_t$ of 0.583 (0.603) at $\tau_l$ and 0.411 (0.415) at $\tau_h$.
The above-mean tokens that $m^-$ selects in incorrect responses have mean (median) $q_t$ of 0.974 (0.998) at $\tau_l$ and 0.956 (0.993) at $\tau_h$.

\subsection{Retained Data and Supervision}
\label{app:cost}

Table~\ref{tab:cost-full} extends Table~\ref{tab:cost} with the retained responses, selected epochs, and optimizer steps.
Supervised tokens count active loss positions through the selected checkpoint, including repeated use across epochs and the positions that $m^-$ selects in incorrect responses.
Relative to training on all candidates, Random supervises 51.2\% fewer tokens and GIST 66.6\% fewer; GIST supervises 31.5\% fewer than Random.
Random and GIST retain the same number of responses but not the same responses, so the per-epoch reduction from masking also reflects response-length differences.

\begin{table}[h]
\centering
\caption{Retained data and supervision on the shared 1,593-prompt pool (Qwen3-0.6B/GSM8K, BOLT).}
\label{tab:cost-full}
\small
\begin{tabular}{lccc}
\toprule
Quantity & All candidates & Random & GIST \\
\midrule
Retained responses              & 50,976 & 12,744 & 12,744 \\
Selected epoch                  & 3 & 5 & 6 \\
Supervised tokens (M)           & 13.596 & 6.641 & 4.546 \\
Supervised tokens per epoch (M) & 4.53 & 1.33 & 0.76 \\
Optimizer steps                 & 4,779 & 1,995 & 2,394 \\
Test Strict / Flex.\ (\%)       & 43.82 / 46.70 & 43.14 / 48.60 & 45.79 / 49.73 \\
\bottomrule
\end{tabular}
\end{table}

\section{Baselines and Ablation Controls}
\label{app:baselines}

All selectors and controls use the shared candidate pool, the 1,593 eligible prompts, and the BOLT objective of Appendix~\ref{app:impl}, unless stated otherwise.

\paragraph{Selection baselines.}
\emph{Random} samples four correct and four incorrect responses per prompt uniformly from the combined temperature pools and supervises all eligible tokens.
\emph{RFT-style} retains all verifier-correct responses.
The HES, PODS, ProFit, ETS, and RSI-S adaptations apply each method's selection rule within this pipeline instead of reproducing its full training algorithm, and each retains eight responses per prompt.
HES scores a response by the summed predictive entropy of its top 0.5\% highest-entropy positions and keeps the four highest-scoring correct responses and four uniformly chosen incorrect ones.
PODS keeps the eight responses that maximize the variance of their binary rewards, which gives four of each class.
Both supervise all eligible tokens.
ProFit, ETS, and RSI-S keep GIST's responses and select tokens in both classes with one shared rule.
ProFit keeps tokens whose frozen-policy probability is at least 0.1 (96.9\% of tokens), ETS keeps the 20\% of tokens with the highest predictive entropy, and RSI-S keeps tokens whose score $1+\log p_t/H_t$ lies in $[-6,0.95]$ (99.3\% of tokens).

\paragraph{Ablation controls (Tables~\ref{tab:components} and~\ref{tab:token-rule}).}
The Mask column of Table~\ref{tab:components} and the token rules of Table~\ref{tab:token-rule} vary the treatment of correct responses, on which the BOLT weight concentrates.
In Table~\ref{tab:components}, row~(a) trains on all 32 candidates of each eligible prompt with all eligible tokens, and row~(b) is the Random selector.
Row~(c) applies GIST's masks to the same responses as Random.
Row~(d) keeps GIST's responses, supervises all eligible tokens of correct responses, and uses $m^-$ for incorrect ones; it is the All tokens entry of Table~\ref{tab:token-rule}.
Row~(e) keeps GIST's responses and masks and sets $w=1$ for every response, which is SFT + GIST.
In the token-rule ablation, \emph{Random} keeps, in each correct response, as many uniformly chosen positions as GIST selects, and \emph{Highest-probability} keeps the same number of highest-probability positions; incorrect responses use $m^-$ in all variants.

\end{document}